%% file: refs.tex
\documentclass[runningheads]{llncs}
\usepackage{graphicx}

\usepackage{tikz}
\usepackage{comment}
\usepackage{amsmath,amssymb} 
\usepackage{color}
\usepackage[misc]{ifsym}
\usepackage{subcaption}
\usepackage{multirow}
\usepackage{array}

\usepackage[accsupp]{axessibility}  

\usepackage{color}

\usepackage{tikz}
\usepackage{subcaption}

\newcommand\blfootnote[1]{%
  \begingroup
  \renewcommand\thefootnote{}\footnote{#1}%
  \addtocounter{footnote}{-1}%
  \endgroup
}
\begin{document}
\pagestyle{headings}
\mainmatter
\def\ECCVSubNumber{1626}  

\title{Personalizing Federated Medical Image Segmentation via Local Calibration} 


\titlerunning{Personalizing Federated Medical Segmentation via Local Calibration}
%

\author{Jiacheng Wang\inst{1}\and
Yueming Jin\inst{2}\and
Liansheng Wang\inst{1}\textsuperscript{(\Letter)}}
\authorrunning{Wang and Jin et al.}
%
\institute{
Department of Computer Science at School of Informatics,\\ Xiamen University, Xiamen, China \\
\email{jiachengw@stu.xmu.edu.cn, lswang@xmu.edu.cn}  \and
Wellcome/EPSRC Centre for Interventional and Surgical Sciences (WEISS) and Department of Computer Science, University College London\\
\email{yueming.jin@ucl.ac.uk}
}
\maketitle

\input{p0_abstract}
\input{p1_introduction}
\input{p2_related_work}
\input{p3_method_new}
\input{p4_experiment}

\input{p5_conclusion}
\section*{Acknowledgement} This work was supported by the Ministry of Science and Technology of the People's Republic of China (2021ZD0201900)(2021ZD0201903).
\clearpage
%
%
\bibliographystyle{splncs04}
\bibliography{refs}
\end{document}

%% file: p0_abstract.tex
\begin{abstract}
Medical image segmentation under federated learning (FL) is a promising direction by allowing multiple clinical sites to collaboratively learn a global model without centralizing datasets.
However, using a single model to adapt to various data distributions from different sites is extremely challenging.
\blfootnote{\textsuperscript{} J. Wang and Y. Jin---Contributed equally.}
\blfootnote{\textsuperscript{} L. Wang---The Corresponding author.}
Personalized FL tackles this issue by only utilizing partial model parameters shared from global server, while keeping the rest to adapt to its own data distribution in the local training of each site. 
However, most existing methods concentrate on the partial parameter splitting, while do not consider the \textit{inter-site in-consistencies} during the local training, which in fact can facilitate the knowledge communication over sites to benefit the model learning for improving the local accuracy.
In this paper, we propose a personalized federated framework with \textbf{L}ocal \textbf{C}alibration (LC-Fed), to leverage the inter-site in-consistencies in both \textit{feature- and prediction- levels} to boost the segmentation.
Concretely, as each local site has its alternative attention on the various features, we first design the contrastive site embedding coupled with channel selection operation to calibrate the encoded features.
Moreover, we propose to exploit the knowledge of prediction-level in-consistency to guide the personalized modeling on the ambiguous regions, e.g., anatomical boundaries. It is achieved by computing a disagreement-aware map to calibrate the prediction.
Effectiveness of our method has been verified on three medical image segmentation tasks with different modalities, where our method consistently shows superior performance to the state-of-the-art personalized FL methods.
Code is available at \url{https://github.com/jcwang123/FedLC}.
\end{abstract}

%% file: p1_introduction.tex
\section{Introduction}
\input{ref_figures/Figure_fl}
As a data-driven approach, deep learning model heavily relies on the data quantities to prompt its efficacy.
Collaborative training using the data across multiple medical sites is increasingly essential for yielding the maximal potential of deep models for medical image segmentation~\cite{zavala2020segmentation,dong2021polyp,sarhan2020utilizing}.
However, it is generally infeasible to accomplish the data communication over multiple sites owing to the privacy protection for patients~\cite{kaissis2020secure}.
Federated Learning (FL)~\cite{konevcny2016federated} has recently received significant research interests from the community~\cite{mcmahan2017communication,rieke2020future,kaissis2020secure}, as it enables the different sites to jointly train a global model with no need to share and centralize the data.
Instead, each local client (e.g., medical site) trains the model from their own data, and the coordinating is achieved by aggregating the model parameters from the local clients to a global server and broadcasting the updated parameters to them.
See a typical and standard federate paradigm in Fig.~\ref{fig:fl}(a), a single global model is generated in the server by averaging the model parameters from the local clients.

Although FL has recently achieved the promising progress in medical image segmentation~\cite{sheller2018multi,li2019privacy,li2020federated,liu2021feddg}, most existing works fall into the standard federate learning paradigm, i.e., concentrating on learning \emph{a single global model} with more robustness and generalization on clients via balanced average weight~\cite{li2019privacy}, image simulation~\cite{li2020federated}, image style transformation~\cite{liu2021feddg}.
However, the single global model can not perform well on all local sites due to the data heterogeneity, where the underlying data sample distribution of local sites could be substantially different from each other~\cite{fallah2020personalized,tan2022towards}. 
The potential model degradation is more severe in the medical scenario, where the difference of scanners, imaging protocols, patient populations bring the high diversity of data distribution. 
In this work, we focus on improving the performance of each local client, by exploiting the data of all clients but learning \emph{a personalized model for each client}, which is highly desired for practical usage yet still underexplored in medical image segmentation tasks.

Personalized FL emerges and establishes a promising approach for improving the quality of each local client model~\cite{wang2019federated,yu2020salvaging,2021personalized,andreux2020siloed,li2021fedbn,collins2021exploiting}.
Among them, the vanilla solution is delivering the global model to local clients back and using their own data to do the model finetuning~\cite{wang2019federated,yu2020salvaging}.
However, these methods adjust the full-dimensional parameters, which may destroy the common representation gained by FL and negatively affect the performance.
%
Recent advances reduce the communication part of local parameters, where only the partial parameters will be sent to the global server for updating (global part) and others (personalized part) are maintained in the local site.
The personalized part can be concluded into two streams.
One is at feature-level such as the high-frequency components of convolutional parameters~\cite{2021personalized} or Batch Normalization layers~\cite{andreux2020siloed,li2021fedbn}.
The other one is at the prediction level, i.e., the prediction head layers~\cite{collins2021exploiting} (as shown in Fig.~\ref{fig:fl}(b)).
However, the two streams still only consider the intra-site information during local training, while they have ignored to exploit the \textit{inter-site inconsistency}. The valuable knowledge from other sites shall be inevitably lost.
Additionally, most existing literature on personalized FL tackles the classification problem, in which the classification models contain much fewer and simpler layers than the segmentation models so that whether they are useful and how to effectively form the personalization for segmentation tasks are still under-explored so far.

In this paper, we propose a personalized federated segmentation framework that is able to unify the personalized feature representation and target prediction through \textbf{L}ocal \textbf{C}alibration, so-called \textbf{LC-Fed} as shown in Fig.~\ref{fig:fl}(c).
The feature- and prediction-level personalization is respectively achieved by the Personalized Channel Selection (PCS) module and the Head Calibration (HC) module.
\begin{itemize}
    \item
    \textbf{PCS} can calibrate the encoded features after standard encoding layers depending on our proposed \textit{contrastive site embedding}, which is unique for each site and inter-site contrastive. Specifically, given a site embedding and the encoded features, as each site pays its personalized attention to the various channels, the PCS module augments the site embedding and yields an attention channel factor to calibrate the features.
    
    \item \textbf{HC} is designed with the insight that, the inter-site in-consistency at prediction-level always implies the most ambiguous areas which demand more concentration in model training. In order to take advantage of this prior knowledge, HC gathers prediction heads from other sites and calculates a \textit{disagreement-aware map} to calibrate its own prediction. 
\end{itemize}

We conduct comparison experiments with several personalized FL methods on three typical medical image segmentation tasks, which are prostate segmentation on T2-weighted MR images (PMR), polyp segmentation of endoscopic images (EndoPolyp), and optic disc/cup segmentation of retinal fundus images (RIF). 
We evaluate the local accuracy of all federated sites and calculate the averaged score to assess the performance. 
Experimental results demonstrate the effectiveness of the proposed method, consistently achieving better segmentation results than the state-of-the-arts.

%% file: ref_figures/Figure_fl.tex
\begin{figure}[!t]
    \centering
    \includegraphics[width=.7\linewidth]{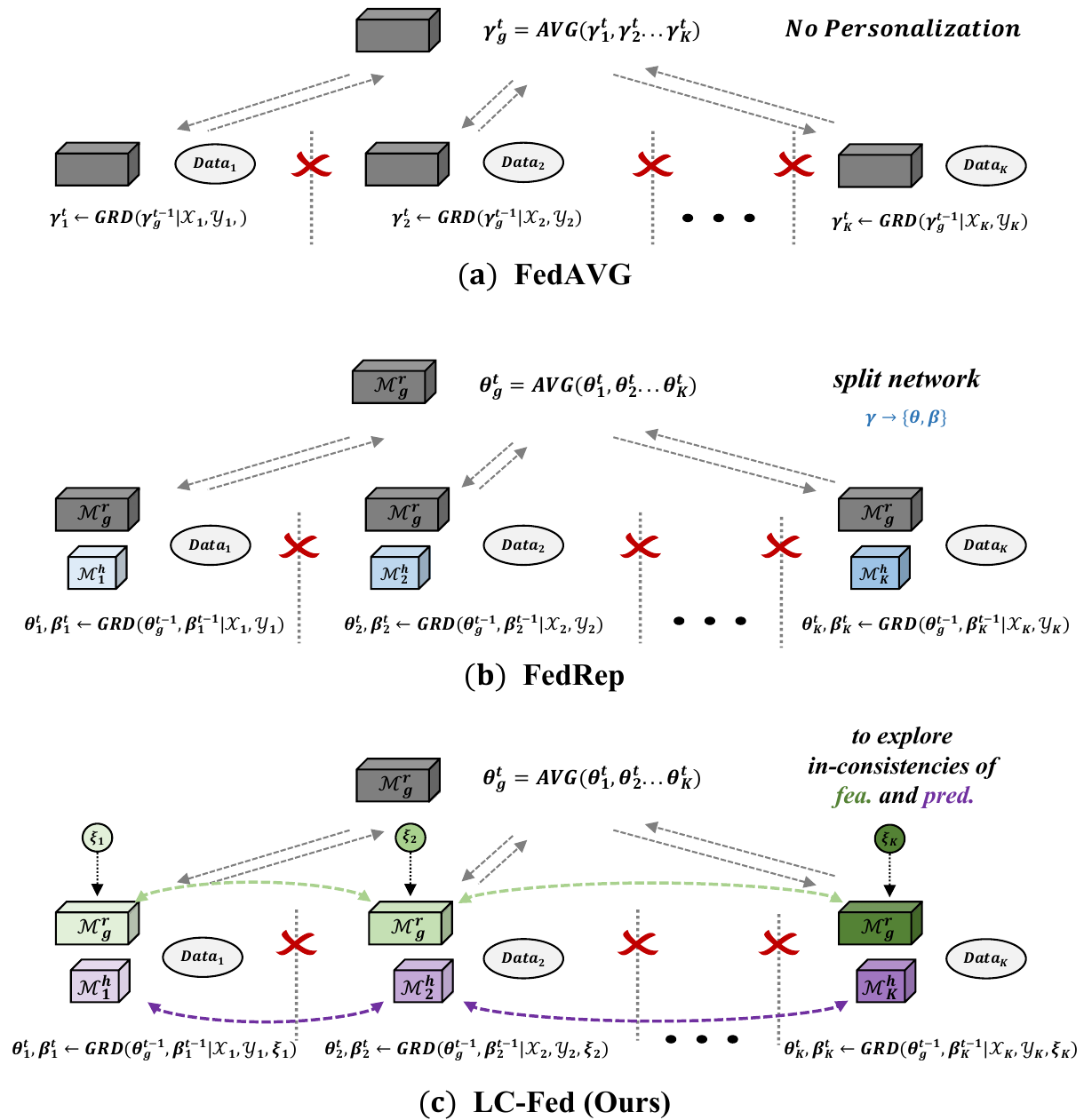}
            \caption{Federated frameworks. Dash lines with arrow denote the parameter communication. 
    (a) FedAVG, the classical federated framework that designs the global model averaging all parameters from local sites.
    (b) FedRep, the latest personalized framework under federation, which splits the model into representation part and head part. The former is updated through global averaging while the latter is trained only using local data.
    (c) Our method LC-Fed, that explores inter-site in-consistencies to calibrate the local learning from both feature- and prediction- levels.}\vspace{-0.2cm}
    \label{fig:fl}
\end{figure}
    

%% file: p2_related_work.tex
\section{Related Work}
\label{sec:related_work}
\subsection{Medical Image Segmentation}
Medical image segmentation aims to predict a certain region from an input image, such as organ at risk from MR image~\cite{tian2018psnet,zavala2020segmentation}, optic disc and cup from retinal fundus image~\cite{thakur2019optic,sarhan2020utilizing}, and polyp from endoscopic image~\cite{fan2020pranet,zhang2021sa,dong2021polyp}.
It contributes a lot to the improvement of clinical efficiency, treatment effect, and quality control.
A large proportion of current research focuses on architecture adjustment to improve representation ability, by attention-like mechanism~\cite{dong2021polyp}, multi-scale feature fusion~\cite{fan2020pranet}, hyper-architecture~\cite{sarhan2020utilizing} and so on.
However, learning powerful representation requires a large amount of data in general, or the performance will meet a serious drop.
To increase the data amount, collecting data over different sites is mostly necessary as each institution has limited patients especially for some rare diseases, while the data communication is sometimes hard to come true due to privacy protection of patients' information.
Hence, instead of improving the architectures, we pay attention to building a federated segmentation framework with no need to centralize data from different sites. 
Also, we make efforts to personalize the federation so that local accuracy can be extensively improved that benefits the local application.

\subsection{Federated Learning}
With the consideration of increasing attention on privacy legislation, federated learning is catching more and more eyes in recent years, particularly in the medical area, which requires no need to centralize data over different sites~\cite{silva2019federated,li2019privacy,rieke2020future,kaissis2020secure}.
The federated learning paradigm is meant to protect patients' privacy and can even achieve competitive performance compared to that of models trained on centralized data~\cite{kaissis2020secure}.
Its workflow can be realised with several different topologies, i.e. centralized server and decentralized sever, but the goal remains the same that is to aggregate knowledge from different sites without data communication~\cite{rieke2020future}.
For example, the most classical one, FedAVG~\cite{mcmahan2017communication}, proposes to average parameters from all local sites at the centralized server.
In the medical area, Sheller et al.~\cite{sheller2018multi} firstly conducts a pilot study to investigate the usefulness of FL in multi-site brain tumor segmentation and recent studies aim to tackle the quantity imbalance over sites~\cite{li2020federated,zhang2021federated} or enhance the generalization ability during federated setting~\cite{liu2021feddg}
In other areas, numerous methods have been introduced to solve the various FL challenges~\cite{kairouz2019advances}, such as reducing communication cost~\cite{mcmahan2017communication,reisizadeh2020fedpaq}, privacy protection~\cite{zhu2020federated}.
However, in whatever areas, when applied for practice, the goal of each participated host is to obtain an accurate model for its local data, while using a single global model to adapt to different data distribution is extremely hard. 
With the motivation to boost the local accuracy as much as possible and ignore the probable performance drop on the unseen domain, we propose to personalize the federated medical image segmentation, in which each site has its own parameters and simultaneously catch knowledge from other sites.

\subsection{Personalized Federated Learning}
The data heterogeneity, differences in data distribution, makes it hard to learn a single global model that can be applied to all sites.
To cope with this issue, personalized federated learning is introduced to personalize the global model uniquely for each participating client in the setup~\cite{kulkarni2020survey}.
Some work treats personalized federated learning as a multi-task learning problem where each site's learning process is a unique task~\cite{dinh2021fedu,marfoq2021federated}.
Other approaches divide the network architecture into shared and personalized layers, where the shared layers are aggregated by FedAVG at the centralized server and the personalized layers are not.
The shared layers could be batch normalization layers~\cite{li2021fedbn}, high-frequency convolution layers~\cite{2021personalized} or prediction head layers~\cite{collins2021exploiting}.
However, these setups are all designed and evaluated for classification tasks while the segmentation models have more complex architecture so the effectiveness has not been verified.
Additionally, these methods have not adequately investigated the inter-site disagreement at both feature- and prediction-level, which is beneficial to concurrently learn the personalization and communication.
To fill this gap, our work introduces a novel personalized FL paradigm considering the inter-site information in the local training and demonstrates the promising performance on medical image segmentation task.

%% file: p3_method_new.tex
\section{Method}
An overall of our personalized federated segmentation framework of LC-Fed is visualized in Fig.~\ref{fig:framework}.
It takes the first attempt to enhance the inter-site communication by exploring inconsistencies between sites in the local training.
We start with introducing the overview of our personalized federation paradigm with local calibration, then describe the two modules in detail at the rest parts.
%
%
\input{ref_figures/Figure_frame}

\subsection{Locally Calibrated Federation Paradigm}
\label{sec:overall}

Denote ($\mathcal{X}$,$\mathcal{Y}$) as the joint image and label space with $K$ sites.
For the $k$-th site, the data samples establish its own data distribution ($x_k$,$y_k$)  $\sim$ $ D_k$, where $x_k \in \mathcal{X}_k$ and $y_k \in \mathcal{Y}_k$.
Instead learning a single shared model, our LC-Fed aims at learning $K$ unique models stylized for $K$ local sites $\{\mathcal{M}_i\}_{i=1}^K$ : $\mathcal{X} \rightarrow \mathcal{Y}$ using the whole data space.
The models hold the same network architecture, containing two components: a base body for representation learning $\mathcal{M}^{r}$ with parameter $\theta$, a prediction head to map the representation to produce predicted values for each class $\mathcal{M}^{h}$  with parameter $\beta$.
Specifically, an U-shape network is exploited as the base body $\mathcal{M}^{r}$, consisting of five encoding-decoding stages. Our feature-level calibration is achieved by injecting the contrastive site embedding $\xi_k$ into the final encoding stage in the local training, to incorporate the stylized information of each local site into the representations from coarse to fine.
Our prediction-level calibration is established within the prediction head $\mathcal{M}^{h}$, which contains two cascaded fully-connected layers, to produce a coarse map and a calibrated segmentation map, respectively.

To learn $K$ unique local models, our LC-Fed alternates between the local site update and global server update on each communication round.
At each federated round $t$, all local sites receive the same parameters $\theta^{t-1}_{g}$ for the base body part from the global server at the last round, while the parameters of the prediction head are initialized from the local training itself, i.e., $\beta^{t-1}_{k}$ in $k$-th local site.
Each site will update the model for its optimal solution using its local data ($\mathcal{X}_k$,$\mathcal{Y}_k$) and the stylized site embedding $\xi_{k}$ as
\begin{equation}
    \theta^{t}_{k},\beta^{t}_{k} \leftarrow \text{GRD} (\theta^{t-1}_{g},\beta^{t-1}_{k}| \mathcal{X}_k, \mathcal{Y}_k, \xi_{k}),
\end{equation}
where $\text{GRD}(.)$ represents the local gradient-based update.
After the update finished in all local sites, the global server then collects the parameters for the representation portion $\theta^{t}$ to update the global model.
We employ the most popular federated averaging algorithm (FedAvg)~\cite{mcmahan2017communication}, which performs an average operation on the local parameters for global model updating, i.e., $\theta^{t}_{g}=\frac{1}{K}\sum_{i=1}^{K}\theta^{t}_{i}.$
Till now, the current federated round finishes and the global server shall deliver the updated representation $\theta^{t}_{g}$ to local sites to turn to the next round.

\subsection{Personalized Channel Selection via Contrastive Site Embedding}
\label{sec:pcs}
The strong representation ability of convolutional layers can be owing to various feature channels to some degree, where each channel represents a unique perspective for the target learning.
Considering the underlying distribution variance of image data between different local sites, it is desirable for each site to pay its personal attention to alternative channels.
With this motivation, we propose a stylized site embedding, coupled with the designed channel selection operation, to calibrate the local sites to pursue their own feature representation modeling with different directions.
Moreover, the simple site embedding design invokes the inter-site communication during the local feature learning, which prompts the inter-site inconsistency in the feature level. 
It is achieved by incorporating the contrastive objective on the site embeddings to incite them to be dissimilar.

Concretely, in the $k$-th local site, the encoded feature of a image data at the $l$-th encoding stage is denoted as $f_k^l \in \mathbb{R}^{C \times \frac{H}{2^{l}} \times \frac{W}{2^{l}}}$, where $C$ is the channel number and $(H,W)$ are the image size and $l=5$ in this work.
To simplify the communication cost, we initialize the site embedding as a one-hot vector, with the length set as the site number $K$: $\xi_{k} \in \mathbb{R}^{K}$.
The $k$-th value is 1 and others are 0.
We further integrate the textural semantics from the feature in the current stage $f_k^l$ to enhance the site embedding.
To achieve this, we first extend the length of site embedding to keep balance with the textural semantics for better training stability, i.e., the updated one $\xi_{k}^* \in \mathbb{R}^C$.
We employ two fully-connected layers, with instance normalization and Relu activation in between, to accomplish the extension.
We then perform the global averaging on each channel of the feature $f_k^l$, to generate a channel descriptor that can represent the abundant textural information and save the computational cost.
The feature concatenation on channel descriptor and $\xi_{k}^*$, followed by a full-connection and a gating sigmoid activation is used for augmenting site embedding by the textural knowledge. 
The augmented site embedding $\hat{\xi_{k}}$ can serve as the attentive factor for selecting feature channels. We use the residual design for reducing the negative effects caused by wrong selection: $f_k^{'} = f_k + f_k \otimes \hat{\xi_{k}}$, where $\otimes$ denotes the pixel-wise multiplication.
$f_k^{'}$ is then fed into the decoder followed by the prediction head to generate the segmentation map.
As the initially injected site embeddings $\xi_{k}$ are mutually independent and different among local clients, they can generate different augmented versions to calibrate the feature representation learning to adapt to their own data distributions and not affect others.

\noindent\textbf{Site-Contrast Regularization.}
To prompt the inter-site inconsistency in the current feature-level calibration, we present the site-contrast regularization to encourage a larger distance between different site embeddings.
%
Taking the $k$-th site's regularization as an example, we sequentially couple $f_k$ with each site embedding in $\{\xi_i\}_{i=1}^{K}$, and feed them into the generator $\mathcal{F}_{cs}$ one by one to obtain a set of augmented version $\{\hat{\xi}_i\}_{i=1}^{K}$, which are used for channel selection.
We \textbf{maximize} distance between $k$-th augmented site embedding $\hat{\xi}_k$ and others as:
\begin{equation}
\label{eq:regularization}
    \mathcal{L}_{con} = -\frac{1}{K-1}\sum\nolimits_{i=1}^{K} |\hat{\xi}_k-\textit{StopGradient}(\hat{\xi}_i)|, ~ \text{s.t.} ~ i\neq k.
\end{equation}
Note that we stop the gradient when augmenting the site embeddings designed for other sites.
To this end, the inter-site feature learning can be facilitated towards different directions by pushing apart the site embeddings.

\begin{figure}[t]
    \centering
    \includegraphics[width=.8\linewidth]{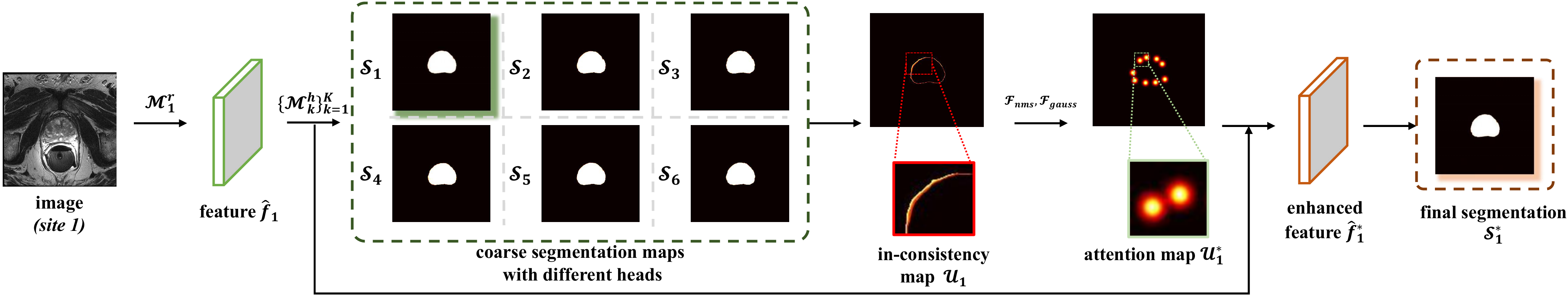}
    \caption{Illustration of head calibration. It starts with calculating an in-consistency map, then encourages the map to concentrate on ambiguous boundaries, and finally utilizes a spatial attending mechanism to empower representations.}
    \label{fig:hc}
    \vspace{-3mm}
\end{figure}

\subsection{Disagreement-aware Head Calibration}
\label{sec:hc}
Previous studies share the base body with the global server while storing the prediction head locally. 
The personalized head is verified to be beneficial for the performance improvement of each site.
In fact, they can be wisely utilized to estimate the inconsistency in prediction level across different sites, and these disagreement regions generally imply the most ambiguous areas that demand more concentration in model training.
For example, the ambiguous boundary of anatomy in Fig.~\ref{fig:hc}.
In this regard, we propose to impose disagreement-aware calibration to incorporate the knowledge of prediction-level inter-site inconsistency, which can guide the model optimization focusing on these challenging regions.

We first extract the inter-site prediction inconsistencies. Considering that the prediction head part $\mathcal{M}^{h}$ is a simple full-connection layer with few parameters, we collect all sites' heads in each local site, that are $\{\mathcal{M}^{h}_i\}_{i=1}^{K}$.
For the $k$-th site, we input the feature $\hat{f}_{k}$ extracted from its base body, into these prediction heads, and result in a set of segmentation maps $\{\mathcal{S}_i\}_{i=1}^{K}$, where each $\mathcal{S}_k \in \mathbb{R}^{N \times H \times W}$ and $N$ is the number of classes (i.e., one for prostate segmentation and two for optic disc/cup segmentation).
Then we construct the disagreement-aware calibration map $\mathcal{U}_{k}$ by calculating the standard deviation along each class channel:
\begin{equation}
    \mathcal{U}_{k}^{c} = \sqrt{\frac{1}{N-1}\sum\nolimits_{i=1}^{K}(\mathcal{S}_{k}^{c}-\mathcal{S}_{i}^{c})^2},
\end{equation}
where $c$ denotes the class channel.
The calibration map keeps the same size as the segmentation map and the pixel value can indicate the disagreement between the other sites and the current local site. 
The larger value suggests more difference that desires more attention in model optimization.

We do not utilize the map directly, instead, we further emphasize the most disagreement regions which can better benefit the model training (see experiments in Sec.~\ref{sec:analysis}).
To do this, we first utilize the Non-Maximum Suppression (NMS) operator $\mathcal{F}_{\text{nms}}$ to identify such regions.
Given each element $u_i$ in $\mathcal{U}_i$, it is only kept when its value is the largest in the surrounding $\delta \times \delta$ values, otherwise, the value is set as 0, where $\delta$ is set as 11 by default, 
We then employ a Gaussian filter $\mathcal{F}_{\text{Gauss}}$ to enlarge the attention area, which is also beneficial for stabilizing the training process~\cite{Ji_2021_MRNet}. The updated calibration map with the same size as the original one is obtained.
We finally perform a pixel-wise multiplication between the updated calibration map and the representation feature $\hat{f}_{k}$, enhanced by a residual design, to incorporate the prediction-level inter-site inconsistency knowledge into the model training: $\hat{f}^{*}_{i} = \mathcal{F}_{\text{Gauss}}(\mathcal{F}_{\text{nms}}(\mathcal{U}_i))*\hat{f}_{i}+\hat{f}_{i}.$
The refined feature is fed into another full-connection layer with Sigmoid activation to predict the segmentation map $\mathcal{S}^{*}_{k}$.

\subsection{Overall Objective in Local Training}
Our model predicts two segmentation maps, i.e., the coarse map $\mathcal{S}_k$ and calibrated map $\mathcal{S}^{*}_k$ in each local site k.
We utilize Dice loss to minimize their difference between the ground-truth segmentation map $\Tilde{\mathcal{S}_k}$, as
\begin{equation}
    \mathcal{L}_{coarse} = 1 - 2*\frac{|\mathcal{S}_k*\Tilde{\mathcal{S}_k}|}{|\mathcal{S}_k|+|\Tilde{\mathcal{S}_k}|} ~,~
    \mathcal{L}_{calib} = 1 - 2*\frac{|\mathcal{S}^{*}_k*\Tilde{\mathcal{S}_k}|}{|\mathcal{S}^{*}_k|+|\Tilde{\mathcal{S}_k}|}.
\end{equation}
Apart from the segmentation loss, the site-contrast regularization enforces our model to yield different site embeddings.
To this end, the overall joint loss for each local site is defined as:
\begin{equation}
    \mathcal{L}_{joint} = \mathcal{L}_{coarse} + \mathcal{L}_{calib} + \lambda \mathcal{L}_{con},
\end{equation}
where $\lambda$ is used to balance the regularization and segmentation. As too large regularization leads to meaningless selection maps, we set it to 0.1 empirically.

%% file: ref_figures/Figure_frame.tex
\begin{figure}[t]
    \centering
    \includegraphics[width=.8\linewidth]{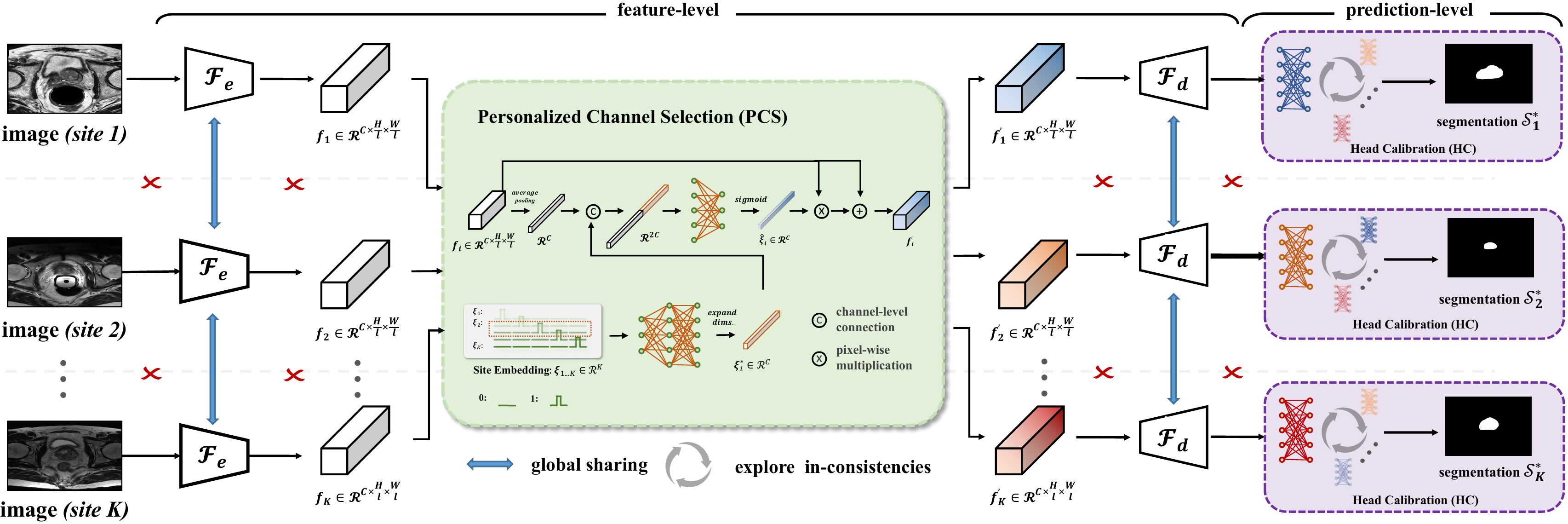}
    \caption{Overview of our personalized federated learning framework with local calibration, LC-Fed. It locally calibrates the features and predictions using the personalized channel selection (PCS) module and head calibration (HC) module.
    In PCS, we propose an unique and contrastive site embedding for each site, through which we calculate a channel selection map to calibrate the feature representation.
    In HC, we gather all sites' prediction heads and measure their in-consistency as the prediction-level disagreement to calibrate segmentation map.}\vspace{-0.2cm}
    \label{fig:framework}
\end{figure}

%% file: p4_experiment.tex
\section{Experiments}
\subsection{Datasets and Evaluation Metrics}
\noindent\textbf{Datasets:} Extensive experiments are conducted to verify the effectiveness of our proposed framework on various medical modalities, including the prostate segmentation from T2-weighted MR images, the polyp segmentation from endoscopic images, and the optic disc/cup segmentation from retinal fundus images.

\begin{itemize}
    \item Prostate MR (\textbf{PMR}) images are collected and labeled from six different public data sources for prostate segmentation~\cite{liu2020ms}. All of them have been re-sampled to the same spacing and center-cropped with the size of $384\times384$. We follow the site division~\cite{liu2021feddg} to divide them into six sites, each of which contains $\{261, 384, 158, 468, 421, 175\}$ slices as well as the labels.
    
    \item Endoscopic polyp (\textbf{EndoPolyp}) images are collected and labeled from four different centers~\cite{bernal2012towards,silva2014toward,bernal2015wm,jha2020kvasir} for polyp segmentation. All the images and annotations are resized to $384\times384$ following the general setting. We follow the latest work~\cite{dong2021PolypPVT} to divide the sites into four parts, each of which contains $\{1000, 380, 196, 612\}$ images and labels.
    
    \item Retinal fundus (\textbf{RIF}) images are collected and labeled from four different clinical centers for optic disc and cup segmentation~\cite{RIM-ONE,orlando2020refuge,sivaswamy2015comprehensive}. We pre-process the data following general setting~\cite{liu2021feddg}, where a $800 \times 800$ disc region in each image is center-cropped uniformly and then resized to $384\times384$. We follow the site division~\cite{liu2021feddg} to split them into four sites, each of which contains $\{101, 159, 400, 400\}$ images as well as the labels.
\end{itemize}

We employ the standard $80\%-20\%$ train-test split widely used in medical vision field for PMR and RIF. The train-test split protocol is slightly different in EndoPolyp where we follow the standard split in the latest work for polyp segmentation~\cite{dong2021PolypPVT}.

\noindent\textbf{Metrics:} We quantitatively evaluate each local site's optimized model on its test data by two commonly-used metrics, including a region-based metric, IoU, and a boundary-based metric, ASSD. The larger IoU and smaller ASSD represent the better segmentation results. 
The averaged scores of all local sites are used for the eventual assessment.

\subsection{Implementation Details}
In the federated learning process, all sites adopt the same hyper-parameters.
$\delta$ in the NMS operator and Gauss filter size are set to 11, and $\lambda$ is set to $0.1$, empirically.
The Adam optimizer with an initial learning rate of $0.0001$ is used to optimize the parameters and the batch size is set as six in all experiments.
Totally, we train the network with 200 rounds as the global model has converged stably and in each federated round, each local site's network is trained with one epoch. 
The whole training process is achieved on the PyTorch platform using one NVIDIA Titan X GPU.

\subsection{Comparison with State-of-the-Arts}
\textbf{Experimental setting.}
We compare our methods to several federated frameworks including the conventional federation, FedAVG~\cite{mcmahan2017communication}, and recent state-of-the-art personalized federation methods, i.e., FedAVG with fine-tuning (FT)~\cite{wang2019federated}, PRR-FL~\cite{2021personalized}, FedBN~\cite{li2021fedbn}, and FedRep~\cite{collins2021exploiting}. 
For implementation, as these methods are originally designed for the image classification task, we try our best to keep their design principle and adapt them to our image segmentation task.
Specifically, we personalize all model parameters in FT; the high-frequency components of convolutional layer parameters in PRR-FL; all the BN layers in FedBN; and the final full-connection layer in FedRep.
We also compare with the baseline setting (Local Train) where each local model is trained using its own data.

\input{ref_tables/Table_comparison_pmr}
\input{ref_tables/Table_comparison_polyp}
\input{ref_tables/Table_comparison_fundus}
\input{ref_tables/Table_ablation_study}

\noindent\textbf{Quantitative Comparison.}
Table~\ref{tab:comparison_pmr} presents the quantitative results of the PMR dataset.
It could be seen that with participating in the federated learning paradigm, the IoU score of Site F largely increases. 
The underlying reason is due to the patient distribution variance, some institutions like Site F have little data to train a powerful deep model if only using their own data.
Thanks to the federation, these institutions obtain a great opportunity to train an employable model by multi-site data collaboration under privacy protection.
We can also see that PRR-FL and FedBN have the interior performance to the basic version (FedAVG), behind which the possible reason is the training collapse.
Straightforward modifying the parameters of convolution or normalization in the personalized part will cause mis-matching of the current feature distribution and desired distribution of convolutional layers.
%
%
In comparison, exploring shared representation and personalizing the prediction layers (FedRep) are slightly useful for the local accuracy improvement.
By investigating the in-consistencies at feature- and prediction levels, our method consistently achieves superior performance over FedRep on all metrics, especially 3.1\% increase on averaged IoU.

Results on EndoPolyp further support the advancement of our method.
Segmentation on this dataset is more challenging because images in each local site only cover a few patients and present limited variance.
Federated learning brings great benefits on this dataset, by leveraging the diverse data from other sites to enhance model learning.
Excitingly, our personalized LC-Fed further attains large result improvement, surpassing FedRep over around 4\% averaged IoU.

As for the RIF dataset, FedAVG shows a close performance compared with ``Local Train".
The reason is that in the RIF dataset, each local site's data is enough to train a satisfactory model. Vanilla federated learning provides limited assistance.
Notably, our method still consistently outperforms FedRep across all the metrics on this dataset.
It demonstrates that, even in the situation that most local sites can provide enough data to train their employable models ($IoU \geq 80\%$), our LC-Fed can still yield great efficacy on personalized federation by considering the inter-site in-consistency.

\input{ref_figures/Figure_quan}
\input{ref_figures/Figure_analysis}
\noindent\textbf{Visual Comparison.}
Fig.~\ref{fig:quan} visually compares the segmentation results on three datasets produced by our method and other personalized methods. 
Apparently, without any federated learning process (LT), the target is hard to be determined in challenging cases (EndoPolyp). 
Using federation (FT) can boost the segmentation in most cases while it fails on some samples. 
The results of PRR-FL and FedBN show huge fluctuation, demonstrating that personalization on the normalization layers and high-frequency convolutional parameters is not stable.
FedRep yields stable and better performance while it sometimes includes the negatives and the determination of boundaries is not precise.
In contrast, our method consistently produces the best segmentation masks.

\subsection{Analytical Ablation Studies}
\label{sec:analysis}
\noindent\textbf{Contribution of key components.}
To prove that the feature- and prediction-level personalization are both useful to improve the local accuracy, we perform an ablation study on all datasets and present the results in Table~\ref{tab:ablation}.
The first row denotes our baseline, FedRep~\cite{collins2021exploiting}.
Comparing it to the second row, we can see that when personalizing the feature representation with our PCS module, the IoU score gains improvement of $1.84\%, 1.01\%, 0.32\%$ on the three datasets.
Shown in the third row, only using the HC module can achieve better results on PMR and RIF datasets.
It is noteworthy that the improvement of using the HC module is smaller than that of PCS, since the segmentation maps are relatively similar without feature calibration, lacking the ability to measure prediction-level in-consistency.
Furthermore, results in the last row from LC-Fed, largely outperform others, demonstrating the complementary advantage of both modules.
Thanks to the feature-level calibration, each site's prediction head adapts to its own feature space so that the HC module can measure prediction-level in-consistency better and the disagreement map contributes to the performance increase.

\noindent\textbf{Contrastive site embedding.}
Previous experiments have verified that the contrastive site embedding is able to strengthen the local accuracy.
To further prove that the performance improvements come from the contrastive comparison instead of extra computation, we conduct the experiment where site embeddings of all local clients are initialized as the same  ($\{\xi_{i}=[1,1,1...1]\}_{i=1}^{K}$) and the site-contrast regularization is removed.
As shown in Figure~\ref{fig:analysis_sub1}, the performance unexpectedly drops significantly on all datasets compared with the baseline.
While using contrastive $\xi$ for each site and adding the site-contrast regularization, our method outperforms the baseline obviously.
It indicates that our PCS module improves the local accuracy by exploring inter-site contrastive features, rather than the extra computation.

\noindent\textbf{Site-contrast regularization factor.}
We further investigate the influence of the hyper-parameter $\lambda$, which is a key factor in the PCS module by controlling the regularization weight in the overall loss.
We vary different $\lambda \in \{-1,-0.1,0,0.1,1\}$ and present results on the PMR dataset in Fig.~\ref{fig:analysis_sub2}.
It shows that the site-contrast regularization can help the module produce more accurate segmentation when comparing $\lambda=\{0\}$ and $\lambda=\{0.1\}$.
In addition, when using opposite regularization ($\lambda=\{-1,-0.1\}$) to pull site embeddings over different sites to be more similar, performances slightly drop as our expectation.
These results demonstrate that repelling site embeddings by our site-contrast regularization is desired and can benefit the feature-level calibration.

\noindent\textbf{NMS operator.}
In the HC module, the NMS operator is utilized to identify the most disagreement regions. Fig.~\ref{fig:analysis_sub3} shows the evaluation results of (not) using the NMS operator on the EndoPolyp dataset. It is seen that the NMS boosts the IoU score from $74.92\%$ into $75.63\%$, indicating that using the NMS operator to filter regions can benefit the model learning.


%% file: ref_tables/Table_comparison_pmr.tex
\begin{table*}[t]
    \caption{
        Quantitative results on PMR dataset. ``FT" means the FedAVG with fine-tuning and $^{*}$ denotes the personalized federation.}
    \resizebox{1.0\textwidth}{!}{%
    \setlength\tabcolsep{1pt}
    \begin{tabular}{c||ccccccc|ccccccc}
        \hline
        \hline
        & \multicolumn{7}{c|}{\textbf{IoU $\uparrow$}}
        & \multicolumn{7}{c}{\textbf{ASSD $\downarrow$}}
        \\
        Sites
        & A & B & C & D & E & F & Avg.
        & A & B & C & D & E & F & Avg.\\
        \hline
        Local Train 
        &77.01&74.15&77.80&76.73&79.67&45.39&71.79
        &0.60&0.93&0.53&0.84&0.86&12.40&2.69
        \\
        \hline
        FedAVG~(\cite{mcmahan2017communication})
        &79.74&80.89&84.91&82.59&83.69&73.27&80.85
        &0.47&0.52&0.32&0.59&0.36&2.02&0.71
        \\  
        FT$^{*}$~(\cite{wang2019federated})
        &81.66&81.51&83.04&80.93&82.09&72.76&80.33
        &0.41&0.59&0.36&0.68&0.39&1.00&0.57 
        \\
        PRR-FL$^{*}$~(\cite{2021personalized})
        &75.10&68.67&79.67&79.71&69.99&51.05&70.70
        &0.79&1.60&0.46&0.79&1.34&12.69&2.95
        \\
        FedBN$^{*}$~(\cite{li2021fedbn})
        &78.91&52.30&77.15&61.75&77.58&64.14&68.64
        &0.54&19.95&0.49&13.80&0.60&1.67&6.17
        \\
        FedRep$^{*}$~(\cite{collins2021exploiting})
        &81.09&81.41&84.70&83.46&82.31&73.81&81.13
        &0.41&0.49&0.32&0.58&0.39&\textbf{0.64}&0.47
        \\
        \hline
        LC-Fed~(Ours)
        &\textbf{85.91}&\textbf{82.27}&\textbf{86.28}&\textbf{85.31}&\textbf{86.08}&\textbf{79.47}&\textbf{84.22}
        &\textbf{0.35}&\textbf{0.48}&\textbf{0.25}&\textbf{0.49}&\textbf{0.32}&0.75&\textbf{0.44}
        \\
        \hline
        \hline
    \end{tabular}
    }
    
    \label{tab:comparison_pmr}
\end{table*}

%% file: ref_tables/Table_comparison_polyp.tex
\begin{table*}[!t]
    \centering
    \scriptsize
    \setlength\tabcolsep{3pt}
    \caption{
        Quantitative results on EndoPolyp dataset. ``FT" means the FedAVG with fine-tuning and $^{*}$ denotes the personalized federation.}
    \begin{tabular}{c||ccccc|ccccc}
        \hline
        \hline
        & \multicolumn{5}{c|}{\textbf{IoU $\uparrow$}}
        & \multicolumn{5}{c}{\textbf{ASSD $\downarrow$}}
        \\
        Sites
        & A & B & C & D & Avg.
        & A & B & C & D & Avg.
        \\
        \hline
        Local Train 
        &48.27&55.26&38.37&62.74&51.16
        &27.89&21.14&36.66&22.62&27.08
        \\
        \hline
        FedAVG~(\cite{mcmahan2017communication})
        &64.56&86.76&61.28&65.93&69.63
        &18.12&2.85&15.33&17.69&13.50
        \\
        FT$^{*}$~(\cite{wang2019federated})
        &65.95&87.45&60.63&69.04&70.77
        &17.43&2.62&16.42&13.65&12.53
        \\  
        PRR-FL$^{*}$~(\cite{2021personalized})
        &15.29&71.69&43.37&73.39&50.93
        &116.87&13.18&31.14&13.15&43.58
        \\
        FedBN$^{*}$~(\cite{li2021fedbn})
        &51.76&78.23&31.21&60.55&55.44
        &30.59&15.16&105.40&28.41&44.89
        \\
        FedRep$^{*}$~(\cite{collins2021exploiting})
        &67.23&\textbf{88.94}&61.17&69.56&71.73
        &16.36&\textbf{2.11}&18.69&16.77&13.48
        \\
        \hline
        LC-Fed~(Ours)
        &\textbf{69.21}&88.51&\textbf{68.10}&\textbf{76.68}&\textbf{75.63}
        &\textbf{15.59}&2.64&\textbf{11.60}&\textbf{12.00}&\textbf{10.46}
        \\
        \hline
        \hline
    \end{tabular}
    
    \label{tab:comparison_fundus}
\end{table*}

%% file: ref_tables/Table_comparison_fundus.tex
\begin{table*}[t]
    \centering
    \setlength\tabcolsep{3pt}
    \scriptsize
    \caption{
        Quantitative results on RIF dataset. ``FT" means the FedAVG with fine-tuning and $^{*}$ denotes the personalized federation.}
    \begin{tabular}{c||ccccc|ccccc}
        \hline
        \hline
        & \multicolumn{5}{c|}{\textbf{IoU $\uparrow$}}
        & \multicolumn{5}{c}{\textbf{ASSD $\downarrow$}}
        \\
        Sites
        & A & B & C & D & Avg.
        & A & B & C & D & Avg.
        \\
        \hline
        Local Train 
        &82.80&78.55&84.80&85.58&82.93
        &5.88&5.12&3.72&2.76&4.37
        \\
        \hline
        FedAVG~(\cite{mcmahan2017communication})
        &84.81&77.88&83.91&84.51&82.77
        &5.19&5.68&3.99&3.00&4.46
        \\  
        FT$^{*}$~(\cite{wang2019federated})
        &85.84&80.21&84.58&85.20&83.96
        &4.62&4.56&3.82&2.85&3.96
        \\
        PRR-FL$^{*}$~(\cite{2021personalized})
        &81.24&78.49&83.75&83.19&81.67
        &7.57&5.09&4.10&3.34&5.02
        \\
        FedBN$^{*}$~(\cite{li2021fedbn})
        &84.70&78.01&85.01&85.13&83.21
        &5.05&5.32&3.65&2.84&4.22
        \\
        FedRep$^{*}$~(\cite{collins2021exploiting})
        &85.33&79.81&83.95&83.53&83.15
        &4.84&4.93&3.95&3.14&4.21
        \\
        \hline
        LC-Fed~(Ours)
        &\textbf{86.33}&\textbf{81.91}&\textbf{85.15}&\textbf{86.81}&\textbf{85.05}
        &\textbf{4.54}&\textbf{4.29}&\textbf{3.62}&\textbf{2.51}&\textbf{3.74}
        \\
        \hline
        \hline
    \end{tabular}
    
    \label{tab:comparison_fundus}
\end{table*}

%% file: ref_tables/Table_ablation_study.tex
\begin{table*}[t]
    \centering
    \setlength\tabcolsep{4pt}
    \scriptsize
    \caption{Ablation study about two components, PCS and HC. We report the averaged IoU and ASSD of three public datasets in this table. The first row without any components denotes the result of FedRep.}
    
    \begin{tabular}{cc||cc|cc|cc}
        \hline
        \hline
        \multirow{2}{*}{PCS}
        & \multirow{2}{*}{HC}
        & \multicolumn{2}{c|}{\textbf{PMR}}
        & \multicolumn{2}{c|}{\textbf{EndoPolyp}}
        & \multicolumn{2}{c}{\textbf{RIF}}\\
        & & IoU $\uparrow$ & ASSD $\downarrow$
        & IoU $\uparrow$ & ASSD $\downarrow$
        & IoU $\uparrow$ & ASSD $\downarrow$\\
        \hline
         & & 81.13 & 0.47 & 72.56 & 11.40 & 83.15 & 3.88  \\
        \checkmark & & 82.97 & 0.44 & 73.57 & 11.23 & 83.47 & 4.23  \\ 
        & \checkmark & 81.65 & 0.46 & 72.21 & 13.10 & 84.10 & 3.94\\ 
        \hline
        \checkmark & \checkmark & \textbf{84.22} & \textbf{0.44} & \textbf{75.63} & \textbf{10.46} & \textbf{85.05} & \textbf{3.74}\\ 
        \hline
        \hline
    \end{tabular}
    \label{tab:ablation}
    \vspace{-3mm}
\end{table*}

%% file: ref_figures/Figure_quan.tex
\begin{figure}[t]
    \centering
    \includegraphics[width=.9\linewidth]{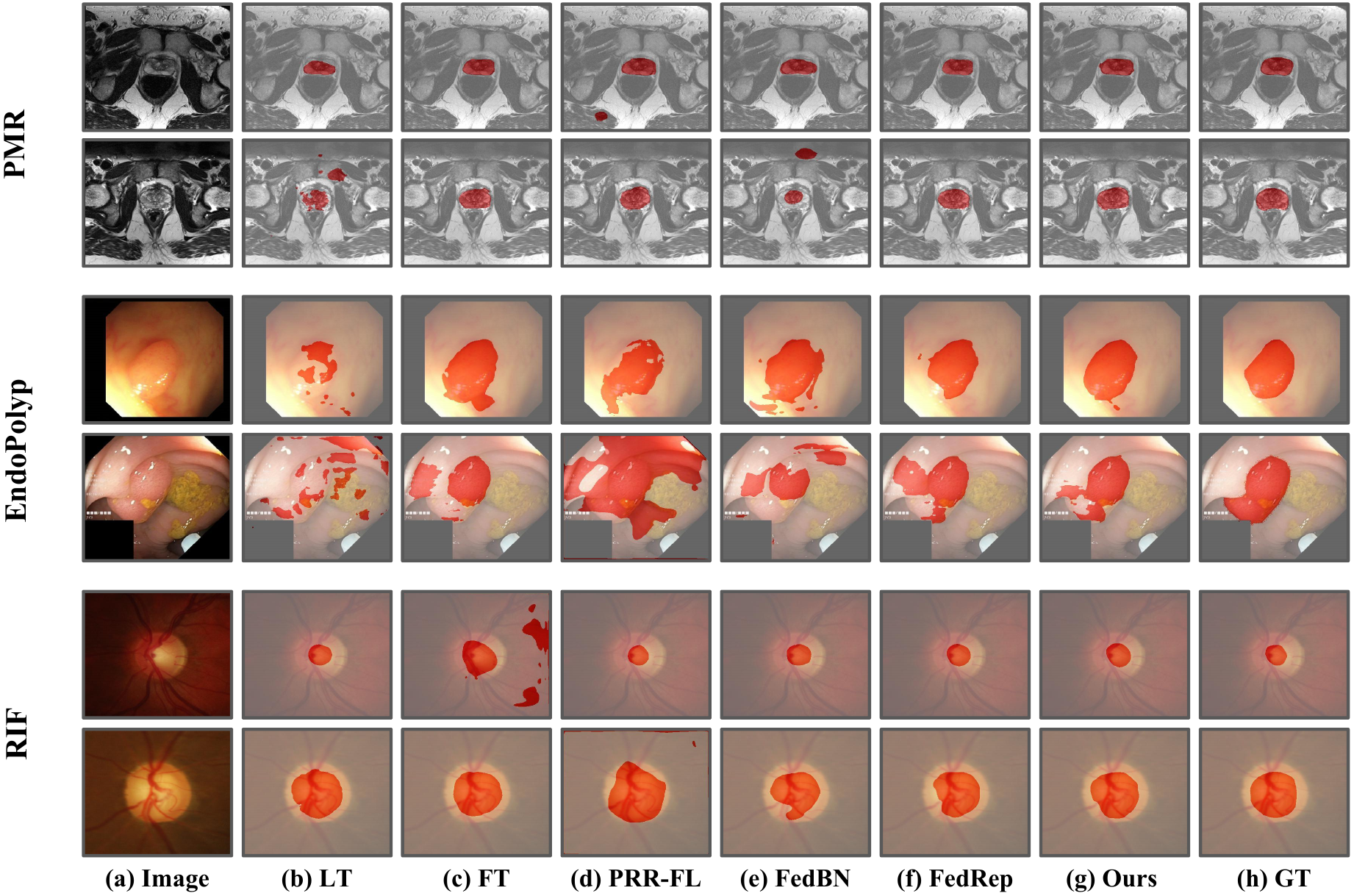}
    \caption{Visualized comparison of the personalized methods on three datasets. From each dataset, we randomly select two samples from different sites to form the visualization.
    (a) Input images from two sites on each dataset;
    (b-g) Segmentation results by model trained with ``Local Train" (LT), FedAVG with fine-tuning (FT)~\cite{mcmahan2017communication}, PRR-FL~\cite{2021personalized}, FedBN~\cite{li2021fedbn}, FedRep~\cite{collins2021exploiting}, and our method LC-Fed;
    (h) Ground truths (denoted as `GT');}
    \label{fig:quan}
    \vspace{-3mm}
\end{figure}

%% file: ref_figures/Figure_analysis.tex
\begin{figure}[th]
    \centering
    \begin{subfigure}{0.3\textwidth}{
        \parbox[][2.2cm][c]{\linewidth}{
            \centering
            \includegraphics[width=\linewidth]{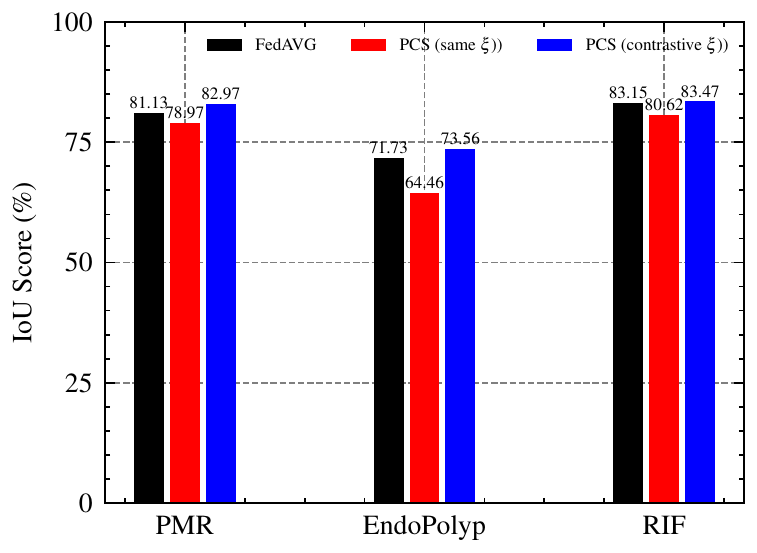}
        }
        \caption{Comparison of employing the same or contrastive site embeddings.}
        \label{fig:analysis_sub1}
    }
    \end{subfigure}\hspace{0.2cm}
    \begin{subfigure}{0.3\textwidth}{
        \parbox[][2.2cm][c]{\linewidth}{
            \centering
            \includegraphics[width=.95\linewidth]{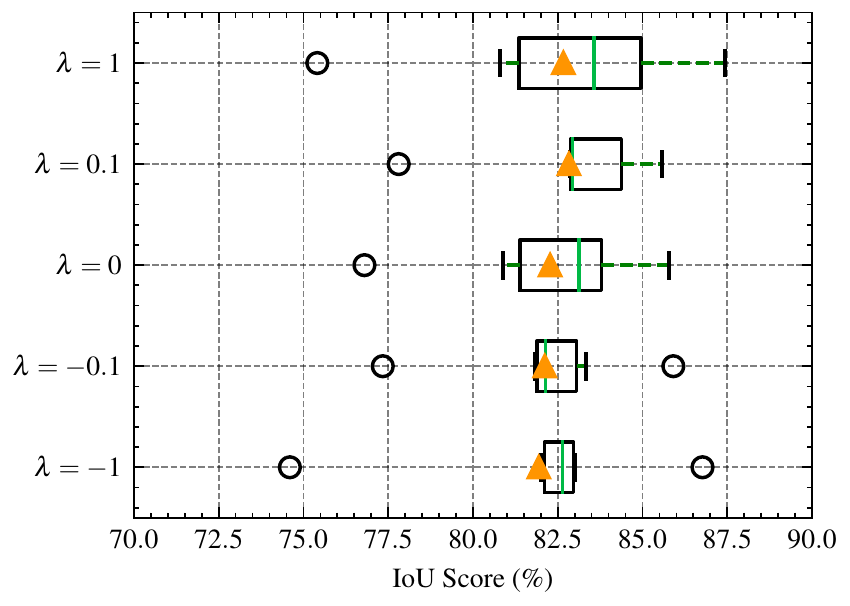}
        }
        \caption{Quantitative analysis using box plots with different $\lambda$ on PMR.}
        \label{fig:analysis_sub2}
    }
    \end{subfigure}\hspace{0.2cm}
    \begin{subfigure}{0.3\textwidth}{
        \parbox[][2.2cm][c]{\linewidth}{
            \centering
            \includegraphics[width=.95\linewidth]{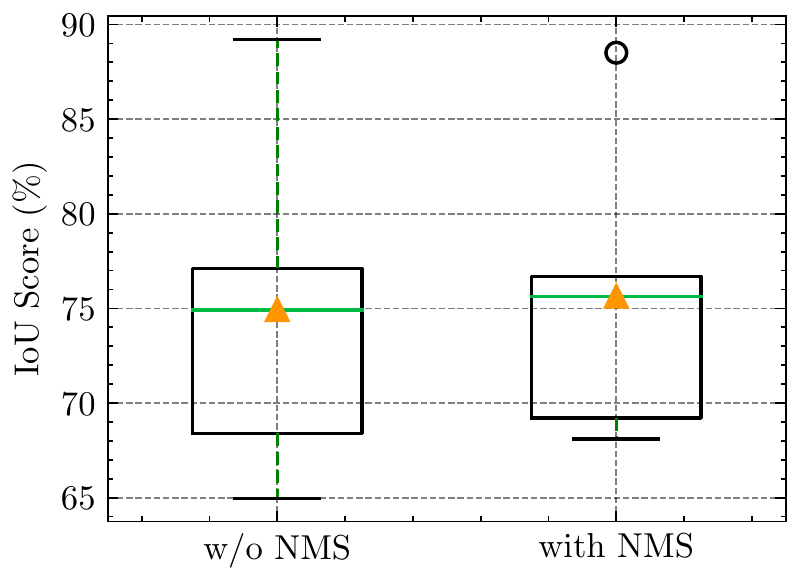}
        }
        \caption{Quantitative analysis about the NMS operator on EndoPolyp.}
        \label{fig:analysis_sub3}
    }
    \end{subfigure}
    \caption{Detailed analysis about the PCS module (a-b) and the HC module (c). The triangles denote the average scores.}\vspace{-3mm}
    \label{fig:analysis_emb}
\end{figure}

%% file: p5_conclusion.tex
\section{Conclusion}
In this paper, we propose to personalize the federated medical image segmentation via unifying the feature- and prediction-level personalization by local calibration. 
The learning paradigm, LC-Fed, is able to calibrate the feature representation and prediction during local training through the Personalized Channel Selection (PCS) module and the Head Calibration (HC) module. 
The PCS module aims to calculate contrastive site embedding for each unique local site and couple it with channel selection operation to pursue the personalized representation modeling.
The HC module is designed to explore the inter-site in-consistency at prediction-level as a disagreement map to calibrate the prediction from coarse to fine.
LC-Fed is evaluated on three public datasets, achieving the best IoU and ASSD on all test sets, compared to previous personalized FL methods.